\documentclass[runningheads]{llncs}
\usepackage{graphicx}
\usepackage{algorithm}
\usepackage{amsmath}
\usepackage{subfigure}
\usepackage[noend]{algpseudocode}
\usepackage{float}
\usepackage{stackengine}
\usepackage{multirow}
\usepackage{booktabs}
\usepackage{arydshln}
\usepackage{tikz}
\usepackage{amssymb}

\usepackage{xcolor}

\newcommand{\xmli}[1]{{\color[rgb]{0.9,0.1,0.3}{}}}
\newcommand{\marwan}[1]{{\color[rgb]{0.1,0.3,0.9}{}}}
\newcommand{\hl}[1]{{\color[rgb]{0.1,0.3,0.9}{}}}

\usepackage[hidelinks,colorlinks=true,linkcolor=blue,citecolor=blue]{hyperref}
\usepackage[capitalize]{cleveref}
\crefname{section}{Sec.}{Secs.}
\Crefname{section}{Section}{Sections}
\Crefname{table}{Table}{Tables}
\crefname{table}{Tab.}{Tabs.}

\begin{document}
%
% \title{Federated Long-tailed Learning via Leveraging Free Lunch Models for Gastrointestinal Image Recognition}
%

\title{Federated Model Aggregation via Self-Supervised Priors for Highly Imbalanced Medical Image Classification}
\titlerunning{Fed-MAS}
% If the paper title is too long for the running head, you can set
% an abbreviated paper title here
%

\author{Marawan Elbatel\inst{1,2} \and
Hualiang Wang\inst{1} \and Robert Martí\inst{2} \and 
Huazhu Fu\inst{3}  \and Xiaomeng Li\inst{1} 
}

\authorrunning{Elbatel et al.} 
% % First names are abbreviated in the running head.
% % If there are more than two authors, 'et al.' is used.
% %
\institute{The Hong Kong University of Science and Technology \and Computer Vision and Robotics Institute, University of Girona \and Institute of High Performance Computing (IHPC), Agency for Science, Technology and Research (A*STAR), Singapore}

% \institute{Princeton University, Princeton NJ 08544, USA \and
% Springer Heidelberg, Tiergartenstr. 17, 69121 Heidelberg, Germany
% \email{lncs@springer.com}\\
% \url{http://www.springer.com/gp/computer-science/lncs} \and
% ABC Institute, Rupert-Karls-University Heidelberg, Heidelberg, Germany\\
% \email{\{abc,lncs\}@uni-heidelberg.de}}
%
\maketitle              

% typeset the header of the contribution
%
\begin{abstract}
In the medical field, federated learning commonly deals with highly imbalanced datasets, including skin lesions and gastrointestinal images. Existing federated methods under highly imbalanced datasets primarily focus on optimizing a global model without incorporating the intra-class variations that can arise in medical imaging due to different populations, findings, and scanners. In this paper, we study the inter-client intra-class variations with publicly available self-supervised auxiliary networks. Specifically, we find that employing a shared auxiliary pre-trained model, like MoCo-V2, locally on every client yields consistent divergence measurements. Based on these findings, we derive a dynamic balanced model aggregation via self-supervised priors (MAS) to guide the global model optimization. Fed-MAS can be utilized with different local learning methods for effective model aggregation toward a highly robust and unbiased global model. Our code is available at \url{https://github.com/xmed-lab/Fed-MAS}.

\end{abstract}

\begin{figure}[t]
    \centering    
\includegraphics[width=0.9\textwidth]{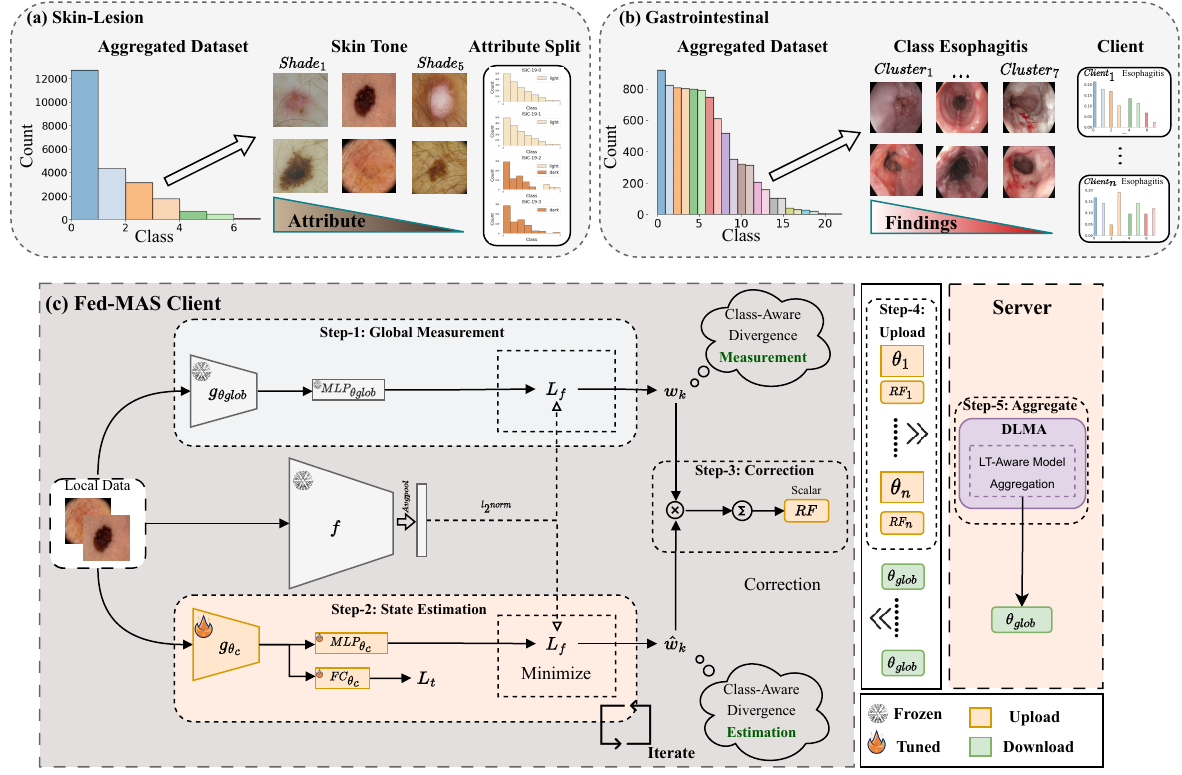}
    \caption{(a) Skin lesion attribute imbalance, (b) Gastrointestinal findings imbalance (Ex: tracheleriazation, varices, leukoplakia). (c) Fed-MAS framework.}    
\label{fig:attrib_imb}
\end{figure}

\section{Introduction}

Federated learning (FL) has emerged as a way to train models with decentralized data while preserving privacy. Due to the inherent nature of data heterogeneity in medical imaging, training in a decentralized manner exhibits performance degradation compared to centralized training. With FedAvg~\cite{McMahan2017CommunicationEfficientLO_fedavg} as the main baseline, multiple works proposed to improve the model's generic performance under data decentralization~\cite{li2021model_moon,LiSZSTS20_fedprox,mendieta2022local_fedalign}. These methods have been successful in achieving positive results, assuming a balanced global data distribution. However, they struggle to address extreme data heterogeneity, especially in highly imbalanced medical datasets. There have been some methods proposed to address the imbalanced setting~\cite{Mu2021FedProcPC,10.1007/978-3-030-87199-4_31_fedirm}. Nevertheless, these methods shared local features among clients, which may raise privacy concerns.

Label distribution skewness has been studied in the context of FL~\cite{pmlr-v162-zhang22p_fedlc}. FedLC~\cite{pmlr-v162-zhang22p_fedlc}, inspired by LDAM~\cite{cao2019learning_LDAM}, showed promising results by adjusting the local client class distribution. Additionally, multiple works proposed to tackle the issue of highly skewed label distribution (i.e. long-tailed) by decoupling the classifier and the feature extractor~\cite{ijcai2022p308_creff,Wicaksana2023FCATL_taming,Chen2022TowardsFL_hualiang_LT}. The rationale behind these methods is rooted in the understanding that the classifier is the bottleneck for majority class bias~\cite{Kang2020Decoupling}. 
For instance, CReFF~\cite{ijcai2022p308_creff} retrained a balanced classifier on the server by leveraging federated features. A notable limitation of classifier re-training is its inability to address the intra-class attribute imbalance. Recently,~\cite{10.1007/978-3-031-20053-3_41_IF_dual} showed that training with imaging data with high attribute imbalance impedes representation learning by exacerbating the intra-class variations. In FL, the issue of intra-class imbalance is critical when dealing with highly imbalanced medical imaging datasets. As depicted in~\cref{fig:attrib_imb} (a), different skin tones can arise across different clients for the same class~\cite{10.1007/978-3-031-16852-9_1_skin_tone_dart}. For gastrointestinal recognition depicted in~\cref{fig:attrib_imb} (b), different findings can arise in different clients for the same class~\cite{Borgli2019HyperKvasirAC}. Hence, the challenge of an unbiased robust global model that takes into account both the attribute and class imbalance still remains. More recently, FedCE~\cite{jiang2023fedce} showed promising results by calculating a fair client contribution estimation in gradient and data space for medical image segmentation; Nevertheless, it relies on local validation samples, which may not adequately represent attribute imbalance and rare diseases in highly imbalanced medical image datasets.

Publicly available pre-trained models, such as MoCo-V2~\cite{He2020MomentumCF_moco} that were trained without any labels using a large set of naturals images, have been utilized with their batch statistics in calculating image priors~\cite{Hatamizadeh2022GradViTGI} and have been utilized with their generalizable representation to improve the performance in highly imbalanced medical imaging tasks~\cite{ding2022free}. In this paper, we leverage these pre-trained models locally to propose Fed-MAS as a novel approach to incorporate the client's local variations with consistent self-supervised priors, estimating client contributing ratios toward an unbiased robust global model.

\section{Methodology} 

\Cref{fig:attrib_imb} shows the overview of our Fed-MAS framework. Each local client is provided with a publicly self-supervised pre-trained model (e.g., MoCo-RN50~\cite{He2020MomentumCF_moco}) that is not involved in the training or communication process of the federated learning framework. Consequently, these pre-trained models do not increase communication costs while ensuring that each client can access the same consistent pre-trained model. With $n$ local clients and one global server, Fed-MAS performs the following steps in each round: (1) Each client receives the global model to measure its global class-aware divergence, $w_k$, and update its local model; (2) Each client trains its local model while estimating its class-aware divergence, $\hat{w}_k$; (3) Each client corrects $\hat{w}_k$ with ${w_k}$ to generate a rescue scalar, $RF$;
(4) Client uploads the parameters of its local model and $RF$ to the server;
(5) The server applies our proposed MAS to aggregate a new model from the parameters of the received client models, weighted by $RF$;

\subsection{Class Aware Global Observation via Self-Supervised Priors}

In highly imbalanced medical image datasets, both extreme class imbalance and inter-client intra-class variations can lead to client drift. Due to the decentralization of data, estimating the global intra-class attribute distribution in medical imaging within the FL framework is a challenge that is yet to be explored.

At the beginning of each round in the FL process, each client receives the model from the global server $\theta_{global}$. We study locally the distance between the distribution of the self-supervised pre-trained model, $f_{\xi}$, and $\theta_{glob}$ over each client's local data.

Given an input image $x$, we feed $x$ to the local feature encoder $g$ to generate a representation $z=g_\theta(x)$. This representation is then fed to an $MLP$ projector to generate a projection $y=MLP_\theta(z)$ in a space comparable with the self-supervised model. From the same discriminative pre-trained model in all clients, we can get a target  representation $y' = f_\xi(x)$, where both $y$ and $y'$ are L2 normalized. We can measure the distribution difference using mean squared error as:
\begin{equation}
    \mathcal{L}^\theta_{f} = 2 - 2\cdot{\langle y,  y' \rangle}
    \cdot
    \label{eq:cosine-loss}
\end{equation}

\begin{figure}[t]
    \centering
\includegraphics[width=\textwidth]{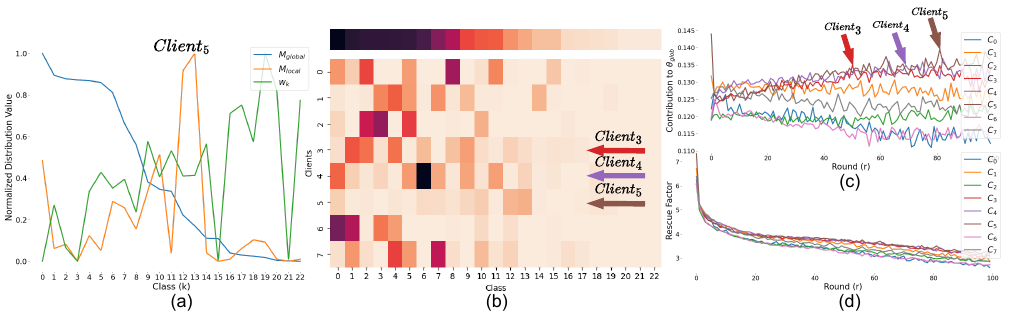}
    \caption{Analysis of MAS on HyperKvasir: (a) The globally aggregated class counts, $M_{global}$, client local count, $M_{local}$, and $w_{k}$ in one round. (b) HyperKvasir non-IID setting, (c) Client's Contribution to $\theta_{glob}$ throughout rounds, (d) Rescue Factor (RF) on different clients throughout rounds} 
    % \caption{$RF_{r,c}$ uses a global measurement, $w_{r,c,k}$, to correct a local estimation, $\hat{w}_{r,c}$.}
\label{fig:RF}
\vspace{-0.5em}
\end{figure}

From~\cref{eq:cosine-loss}, we can generate a class-aware distance for class $k$ with $M_{k}$ total samples as:
\begin{equation}
   \mathcal{L}^\theta_{k} =\frac{1}{M_{k}}\sum\limits_{i = 1}^{M_{k}}  \mathcal{L}_{f}^\theta(x_{k,i}).
   \label{eq:loss_summation_over_theta}
\end{equation}
We define $w_k  = \mathcal{L}_{k}^{\theta_{glob}}$. 
The factor $w_k$ can help to capture the distance in distribution between the global server and the self-supervised model on each client's local data. This divergence can provide insights into the sensitivity of the global model, $\theta_{glob}$, in effectively capturing the specific class attribute in each client's local data. 
A high $w_{k}$ indicates the failure of $\theta_{glob}$ in capturing a local class $k$. In~\cref{fig:RF} (a), we can see that $w_{k}$ is inversely proportional to the global class distribution, even if the local client distribution is not necessarily the same.

\subsection{State Estimation via Knowledge Distillation}
While $w_k$ provides class-aware global divergence measurement with the same consistent local frozen self-supervised model, a client receives the global model, $\theta_{glob}$, and takes subsequent optimization steps for $E$ local epochs with uncertainty to generate $\theta'_{c}$. Hence, the client's drift from the global model is hideous after its uncertain optimization.

With a running average, a client can provide a class-aware divergence likelihood $\hat{w}_{k}$, where $\hat{w}_{k} =\sum\limits_{e = 1}^{E}\mathcal{L}^{\theta'_{c}}_{k}$. The factor $\hat{w}_{k}$ can help to capture how far the client drifted from $f_{\xi}$ since the global measurement, $w_{k}$, was taken. A client can then correct this estimation, $\hat{w}_{k}$, with the global observation, $w_{k}$, to generate a posterior rescue factor, $RF$, in every round.
\begin{equation}
    RF=  \sum\limits_{k=1}^{K}  w_{k} \hat{w}_{k} . \label{eq:rescue_factor}
\end{equation}

A higher $RF$ indicates that the client has information that the global model has not appropriately captured.

To train the projector ${MLP_\theta(\cdot)}$, we propose to minimize~\cref{eq:cosine-loss} along with the local balanced risk minimization~\cite{Ren2020balms} to 
minimize a total loss $\mathcal{L}_{total}$ concerning $\theta$ only as: 
\begin{equation}
\mathcal{L}_{total} = \mathcal{L}_{sup}+ \lambda_{f} ~\mathcal{L}_{f},
\label{eq:loss_total}
\end{equation}
where $\mathcal{L}_{sup}$ refers to the original supervised loss and $\lambda_{f}$ as a weighting factor. 
This can be seen as restricting the client optimization direction. However,
the self-supervised model ensures clients align with a common reference distribution and possess implicit regularization capabilities for minority classes through generalizable features~\cite{Elbatel2023FoProKDFP}.

\subsection{Model Aggregation via Self-Supervised Posteriors}
\begin{figure}[t]
    \centering
\includegraphics[width=\textwidth]{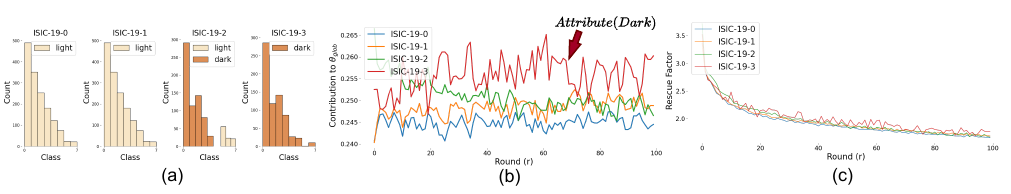}
    \caption{(a) ISIC-FL Attribute Split, (b) Client's Contribution to $\theta_{glob}$ throughout rounds, (c) Rescue Factor (RF) on different clients throughout rounds} 
\label{fig:RF_isic}
\vspace{-0.5em}
\end{figure}

Inspired by the fact that client-specific models should contribute more to the global server to capture local variance, we propose a novel model aggregation via the corrected self-supervised posteriors (MAS) \marwan{should the corrected value be posterior not prior?}. We use our proposed $RF$ to indicate client-specific models that should contribute more to the global model than client-generic models to capture their attribute-class variations. While our proposed $RF$ can be used for biased client selection~\cite{loss_selection}, we use it to aggregate a global model. Instead of aggregating based on the weighted samples as in FedAvg~\cite{McMahan2017CommunicationEfficientLO_fedavg}, we propose to weight the global model, $\theta_{glob}$, based on the $RF$ value as follows:
\begin{equation}
    \bar{RF}_c ={\frac{RF_{c}}{\sum\limits_{j}RF_{j}}} \; \text{, and} \; 
    {\theta^{r+1}_{glob} = \sum\limits_{c = 1}^C  \bar{RF}_c {\theta'_{c}}}.
    \label{eq:tempavg} 
\end{equation}
% \vspace{-5pt}
% \begin{equation}
%     \theta^{r+1}_{glob} = \sum\limits_{c = 1}^C{\frac{RF_{c}}{\sum\limits_{j}RF_{j}}  {\theta_{c}}}
%     \label{eq:tempavg}
% \end{equation}
For instance, Client 3,4,5 in~\cref{fig:RF} (b) have mostly minority classes and contribute the most to $\theta_{glob}$ in~\cref{fig:RF} (c). Morever, in~\cref{fig:RF_isic} (a) Client ISIC-3 have mostly underrepresented attribute and contributes the most in ~\cref{fig:RF_isic} (b). Additionally, we show in~\cref{fig:RF} (d) and ~\cref{fig:RF_isic} (c) that the rescue factor for all clients is decreasing throughout rounds. This highlights the ability of MAS to accommodate different clients. (See~\cref{algo:fedfree} in Appendix).

\section{Experiments}

\noindent{\textbf{Dataset.}}~\textbf{HyperKvasir}~\cite{Borgli2019HyperKvasirAC} is a long-tailed (LT) dataset of 10,662 gastrointestinal tract images with 23 classes from different anatomical and pathological landmarks and findings. We divide the 23 classes into Head ($\textgreater$ 700 images per class), Medium (70 $\sim$ 700 images per class), and Tail ($\textless$ 70 images per class) with respect to their class counts. Additionally, we partition the data across eight clients with IID (similar label distributions) and non-IID (heterogeneous partition with Dirichlet distribution). \textbf{ISIC}~\cite{Combalia2019BCN20000DL_isic2019} is a highly imbalanced dataset of skin lesion images with 8 classes that exhibits skin-tone attribute imbalance~\cite{10.1007/978-3-031-16852-9_1_skin_tone_dart}. For instance, melanoma incidence is lower in quantity and higher in mortality rates in black patients than in others~\cite{Collins2011RacialDI}. We partition the dataset on four clients based on two attributes, light and dark skin tones, with~\cite{10.1007/978-3-031-16852-9_1_skin_tone_dart} labeling. Additionally, we split the data between the four clients for training, validation, and testing with 70\%, 15\%, and 15\%, respectively. We also benchmark Fed-MAS over Flamby-ISIC split~\cite{NEURIPS2022_232eee8e_flamby} with six different hospitals with stratified 5-fold cross-validation.

% \noindent \textbf{ISIC Attribute Split} To investigate the inter-client intra-class variations within a specific attribute, we conduct our study on the ISIC dataset using the skin color attribute. This choice aims to replicate real-world attributes and their distributions.

\noindent\textbf{Implementation Details.}
For both datasets, we use resnet-18~\cite{7780459_resnet} as the local target model. For the long-tailed HyperKvasir dataset, we employ an SGD optimizer and a cosine annealing scheduler~\cite{loshchilov2017sgdr_cosine} with a maximum learning rate of 0.1. For ISIC, we employ Adam optimizer with the 3e-4 learning rate. Additionally, we employ balanced risk minimization~\cite{Ren2020balms} and train methods for 200 communication rounds with 10 local epochs. We set $\lambda_{f}$ to 3 and provide an ablation in~\cref{apx:tbl:free_ablation} in Appendix.

\begin{table}[t]
\centering
\caption{Comparison with other methods on HyperKvasir Dataset. 
%We comprehensively evaluate different LT methods in FL. 
All clients are initialized with ImageNet pre-trained weights; each result is averaged over five runs.}
\label{tbl:sota}
\resizebox{1.0\columnwidth}{!}{
\begin{tabular}{l| c c c |cc | c c c| cc}
% \hline
\hline
     &\multicolumn{5}{c|}{IID} & \multicolumn{5}{c}{non-IID $Dir(\alpha=0.5$)} \\ \hline
   Methods & Head & Medium & Tail & \textbf{All}&\textbf{B-acc} &
   Head & Medium & Tail & \textbf{All}& \textbf{B-acc}\\
%   \hline
% \toprule
   \hline
  % \multicolumn{10}{c}{Centralized Learning (CL)} \\
  %    \hline
% Cross-Entropy & 93.8 $\pm$ 1.3 & 73.9 $\pm$ 1.8& 4.59 $\pm$ 1.3 & 
% 57.47  $\pm$ 0.3
% &
% 58.58 $\pm$ 0.7&
% -$\pm$- & - $\pm$ - & - $\pm$ - & - $\pm$ -\\
  
% LDAM-DRW~\cite{cao2019learning_LDAM}& 90.2 $\pm$ 2.5 & 71.6 $\pm$ 4.1& 21.7 $\pm$ 10 &
% 61.22 $\pm$ 1.9
% & 62.09 $\pm$ 1.7
%    &-$\pm$- & - $\pm$ - & - $\pm$ - & - $\pm$ -\\

% BSM~\cite{Ren2020balms} & 90.5 $\pm$ 3.1 & 64.8 $\pm$ 8.7& 32.9 $\pm$ 8.3 &
% 62.82  $\pm$ 2.0
% & 62.82 $\pm$ 2.1 &
% -$\pm$- & - $\pm$ - & - $\pm$ - & - $\pm$ -\\

%  \textbf{Ours}-$MLP_{s_\theta}$ & 89.8 $\pm$ 1.4 & 68.6 $\pm$ 2.1& 29.3 $\pm$ 5.7 & 
%  62.64 $\pm$ 1.3
% &
% 63.15 $\pm$ 1.2 &
% -$\pm$- & - $\pm$ - & - $\pm$ - & - $\pm$ -\\

% \textbf{Ours}-$MLP_{g_\theta}$ & 86.9 $\pm$ 4.1 & 64.9 $\pm$ 6.6& 36.6 $\pm$ 5.7&
% 62.82 $\pm$ 1.1
% & 63.01 $\pm$ 0.9 &
% -$\pm$- & - $\pm$ - & - $\pm$ - & - $\pm$ -\\

    \hline
     \multicolumn{11}{c}{Federated Learning Methods (FL-Methods)} \\
    \hline
FedAvg~\cite{McMahan2017CommunicationEfficientLO_fedavg} &94.1 $\pm$ 1.3 & 72.9 $\pm$ 1.3& 3.1 $\pm$ 0.9 & 56.69 $\pm$ 0.6& 58.1 $\pm$ 0.6&

86.2 $\pm$ 2.7 & 70.3 $\pm$ 0.5& 8.0 $\pm$ 1.2 & 54.83 $\pm$ 1.0& 56.17 $\pm$ 0.9  \\

FedProx~\cite{LiSZSTS20_fedprox} & 94.6 $\pm$ 0.4 & 72.1 $\pm$ 0.2& 3.0 $\pm$ 1.2 & 56.58 $\pm$ 0.4& 57.93 $\pm$ 0.4 &
88.1 $\pm$ 2.2 & 73.1 $\pm$ 2.7& 3.6 $\pm$ 2.5 & 54.93 $\pm$ 1.3& 56.51 $\pm$ 1.3 
    \\
    MOON~\cite{li2021model_moon} & 94.7 $\pm$ 0.7 & 74.6 $\pm$ 0.4& 4.0 $\pm$ 1.8 & 57.77 $\pm$ 0.6& 59.23 $\pm$ 0.6 &
84.4 $\pm$ 3.6 & 73.1 $\pm$ 1.6& 5.5 $\pm$ 2.1 & 54.3 $\pm$ 1.2& 55.93 $\pm$ 1.1 
\\
    \hline
   
  \multicolumn{11}{c}{LT-integerated FL Methods} \\
  \hline 

LDAM-FL~\cite{cao2019learning_LDAM}&95.4 $\pm$ 0.5 & 72.2 $\pm$ 1.1& 5.7 $\pm$ 3.9 & 57.77 $\pm$ 1.4& 59.03 $\pm$ 1.3 &

86.9 $\pm$ 2.8 & 70.9 $\pm$ 1.2& 4.7 $\pm$ 4.6 & 54.16 $\pm$ 1.4& 55.61 $\pm$ 1.4 \\

BSM-FL~\cite{Ren2020balms} & 93.2 $\pm$ 1.5 & 74.6 $\pm$ 2.6& 9.1 $\pm$ 3.7 & 58.92 $\pm$ 0.6& 60.28 $\pm$ 0.7 &
89.6 $\pm$ 3.9 & 68.7 $\pm$ 3.0& 16.4 $\pm$ 5.4 & 58.24 $\pm$ 1.2& 59.15 $\pm$ 1.3 
\\
   \hline
        \multicolumn{11}{c}{Label-Skew FL Methods} 
        \\
            \hline
         CReFF~\cite{ijcai2022p308_creff} &95.1 $\pm$ 0.8 & 72.0 $\pm$ 1.5& 2.6 $\pm$ 1.8 & 56.53 $\pm$ 1.4& 57.88 $\pm$ 1.4 &

89.3 $\pm$ 0.7 & 70.1 $\pm$ 1.6& 9.0 $\pm$ 4.5 & 56.12 $\pm$ 1.3& 57.34 $\pm$ 1.2 
\\
 FedLC~\cite{pmlr-v162-zhang22p_fedlc} &96.5 $\pm$ 0.4 & 75.3 $\pm$ 2.5& 7.4 $\pm$ 5.5 & \underline{59.73 $\pm$ 1.8}& \underline{61.08 $\pm$ 1.7} &
95.8 $\pm$ 0.6 & 73.1 $\pm$ 2.4& 6.6 $\pm$ 4.1 & \underline{58.51 $\pm$ 1.5}& \underline{59.78 $\pm$ 1.5} 
\\

\hline
\textbf{Fed-Mas (ours)} & 94.3 $\pm$ 1.2 & 72.9 $\pm$ 1.0& 15.9 $\pm$ 2.7 & 
\textbf{61.05 $\pm$ 0.3}&
\textbf{62.08 $\pm$ 0.2} &
93.0 $\pm$ 0.9 & 72.5 $\pm$ 2.6& 16.2 $\pm$ 1.3 &
\textbf{60.57$\pm$ 1.1}&
\textbf{61.61 $\pm$ 1.0} 
   \\
\hline
\end{tabular}}
\vspace{-1em}
\end{table}

\noindent \textbf{Evaluation Metrics.} We evaluate the model performance of the global model in this paper. To assess the unequal treatment of each class in HyperKvasir, we report the top-1 accuracy on shot-based division (head, medium, tail) and their average results denoted as ``All'' as existing works~\cite{10.1007/978-3-031-16437-8_44_monash_sampling_LT}. Following prior work~\cite{10.1007/978-3-030-87240-3_31_balanced_mixup,10.1007/978-3-031-17027-0_3_chest_LT_benchmark,reinke2022metrics}, we also report the Balanced Accuracy ``B-Acc'', which calculates the average per-class accuracy and is resistant to class imbalance. As the test set of HyperKvasir contains only 12 classes, we follow previous work~\cite{10.1007/978-3-030-87240-3_31_balanced_mixup} to assess the model performance with a stratified 5-fold cross-validation. To evaluate the performance of attributes in ISIC-FL, we report the ``B-Acc'' separately for each attribute (``Light'', ``Dark'') and the average of these scores ``Avg''. Additionally, we report the overall ``B-Acc'' across all attributes and distributions.

\subsection{Performance on the HyperKvasir}

We compare our methods with FL-methods~\cite{McMahan2017CommunicationEfficientLO_fedavg,LiSZSTS20_fedprox,li2021model_moon}, LT-integrated FL methods~\cite{cao2019learning_LDAM,Ren2020balms}, and label-skew FL methods~\cite{pmlr-v162-zhang22p_fedlc,ijcai2022p308_creff}

\noindent \textbf{FL-Methods}~\cite{McMahan2017CommunicationEfficientLO_fedavg,LiSZSTS20_fedprox,li2021model_moon}. 
One simple solution for federated learning with highly imbalanced medical data is to apply existing FL methods to our setting directly. 
To this end, we compare our methods with state-of-the-art FL methods, including FedAvg~\cite{McMahan2017CommunicationEfficientLO_fedavg}, FedProx~\cite{LiSZSTS20_fedprox}, and MOON~\cite{li2021model_moon}, under the same setting. As shown in~\Cref{tbl:sota}, we find that our method outperforms the best existing FL method MOON by 2.85\% and 5.68\% on ``B-acc'' in both IID and non-IID settings, respectively. Notably, our Fed-MAS achieves similar results with MOON~\cite{li2021model_moon} on the ``Head'' while reaching large improvements on the ``Tail'' (11.9\% on iid and 10.71\% on non-iid), showing that our Fed-MAS can tackle LT distribution under FL more effectively. 
The limited results could be attributed to the use of local empirical risk minimization in MOON~\cite{li2021model_moon}. However, even when we applied a balanced risk minimization~\cite{Ren2020balms} in MOON, our method still outperformed it (60.69\% vs. 62.08\% on ``B-acc'' for IID); see results in~\Cref{apx:tbl:FL_BRM} in Appendix.

\noindent \textbf{LT integrated FL methods}~\cite{cao2019learning_LDAM,Ren2020balms}. To design FL methods for local clients with long-tailed distribution, a straightforward idea is to directly use LT methods in each local client and then use an FL framework such as FedAvg to obtain the final results. In this regard, we implement LDAM-DRW~\cite{cao2019learning_LDAM} and BSM~\cite{Ren2020balms} into the FedAvg framework and rename them as LDAM-FL and BSM-FL respectively.
From Table~\ref{tbl:sota}, we can notice the LT methods utilizing  an FL framework have produced limited results on ``Tail'' primarily due to the extreme client drifting phenomenon.
Please note that Fed-MAS does not focus on designing any specific long-tailed training for each local client. Instead, MAS enables the global server to effectively aggregate the model parameters from long-tailed distributed local clients.
As a result, our Fed-MAS can successfully capture the ``Tail'' with a 6.84\%  accuracy gain on IID with lower variance than the best-performing LT method BSM-FL~\cite{Ren2020balms}. Notably, our method consistently outperforms the best-performing LT method on the ``B-acc'' with a lower variance (improvement of 1.8\% on IID and 2.46\% on non-IID).

\noindent \textbf{Label-Skew FL} We compare our method with the state-of-the-art label-skew FL method, FedLC~\cite{pmlr-v162-zhang22p_fedlc}, and the highly labeled skew (i.e. long-tailed) FL method, CReFF~\cite{ijcai2022p308_creff}. CReFF, as proposed by~\cite{ijcai2022p308_creff}, involves a method of re-training the classifier by utilizing learnable features on the server at each communication round, holding an equal treatment of all clients' models. However, this technique fails to accommodate inter-client intra-class variations which could arise.
From~\Cref{tbl:sota}, we can notice that FedAvg with local LT such as BSM-FL~\cite{Ren2020balms} can outperform CReFF~\cite{ijcai2022p308_creff} on the HyperKvasir dataset in both IID and non-IDD by 2.4\% and 1.8\% on ``B-acc'', respectively. 
Our comparative analysis illustrates that Fed-MAS consistently outperforms CReFF in both IID and non-IID by 4.2\% and 4.27\% on ``B-acc'', respectively, by incorporating the client's local variations with MAS. FedLC~\cite{pmlr-v162-zhang22p_fedlc} proposes a loss function to address label distribution skewness by locally calibrating logits and reducing local bias in learning. Their modification yields compelling performance. Nevertheless, our method surpasses them in both IID and non-IID, achieving improvements of 1.0\% and 1.83\% on ``B-Acc'', respectively. Remarkably, our method effectively captures the tail classes with reduced variance in both IID and non-IID, exhibiting improvements of 8.5\% and 9.6\%, respectively, while experiencing only a minor drop in performance for the head classes (96.5\% vs 94.3\% for IID and 95.8\% vs 93.0\% for non-IID).

% \noindent \textbf{Fed-LT methods}~\cite{ijcai2022p308_creff} We compare our method with the state-of-the-art Fed-LT method CReFF~\cite{ijcai2022p308_creff}.

% drifting issues with our proposed FLKD (Fast and Convenient Local Model) and DLMA (Robust Estimation for the Global Model) innovations. 

\begin{table}[t!]
\centering
\caption{Ablation of minimizing~\cref{eq:cosine-loss} (KD) and MAS on HyperKvasir non-IID}
\setlength{\tabcolsep}{12pt}
\resizebox{0.8\columnwidth}{!}{
\begin{tabular}{c|c|c|ccc}
\toprule
\multirow{2}{*}{} &\multirow{2}{*}{KD}& \multirow{2}{*}{MAS} & \multicolumn{3}{c}{Metrics}    \\  
\cline{4-6}& & & All (\%)&B-acc (\%) &p-value
   \\ 
   \hline
BSM-FL~\cite{Ren2020balms} (Baseline) & $\times$&$\times$
&58.24 $\pm$ 1.2&
59.15 $\pm$ 1.3 & \textemdash
\\
~\cite{Ren2020balms} w/ KD & \checkmark&$\times$
&59.26 $\pm$ 1.2
& 60.19 $\pm$ 1.1& $<$0.001
\\
\textbf{Fed-MAS}&\checkmark& \checkmark 
&\textbf{60.57 $\pm$ 1.1}
&\textbf{61.61 $\pm$ 1.0} & $<$0.001
\\
\bottomrule
\end{tabular}}
\label{tbl:ablation_dlma_flkd}
% \vspace{-0.5em}
\end{table}

\noindent \textbf{Effectiveness of KD and MAS} 
As shown in~\Cref{tbl:ablation_dlma_flkd}, minimizing~\cref{eq:cosine-loss} (KD) can enhance the ``All'' and ``B-Acc'' via 1.02\% and 1.04\% due to the implicit regularization of MoCo-V2 on the tail classes for extreme imbalance datasets. 
With both KD and MAS, the performance is further improved to the best via 2.33\% and 2.46\% on ``All'' and ``B-Acc'', respectively. MAS utilizes unbiased frozen generalizable representations to incorporate the inter-client intra-class characteristics in FL and combine them with the drifting belief. This combination helps in capturing client-specific models in the aggregation step.

\begin{table}[t]
  \centering
  \caption{Experimental Results on ISIC-FL. Results are averaged over 5 folds.}
  \setlength{\tabcolsep}{12pt}
  \label{tbl:sota_results_isic_fl}
  \resizebox{0.9\columnwidth}{!}{
  \begin{tabular}{l|cc|cc|c}
    \hline
    \multirow{3}{*}{Method}
     & \multicolumn{4}{c|}{\textbf{Attribute Setting (ours)}} & {Flamby-ISIC~\cite{NEURIPS2022_232eee8e_flamby}}
     \\ \cline{2-6}
     & Light &Dark&Avg&B-Acc &B-Acc
     \\
    \hline
    
%   &\multicolumn{5}{c}{w/o Weight Initialization}
%     \\
% \hline

% FedLC~\cite{pmlr-v162-zhang22p_fedlc} & 59.53 $\pm$ 2.7&54.21 $\pm$ 4.3&56.87 $\pm$ 2.2 &59.13 $\pm$ 2.4 
% & 61.26 $\pm$ 3.0
% \\
% BSM-FL~\cite{Ren2020balms} (Baseline) 
% &62.04 $\pm$ 1.7&59.59 $\pm$ 8.4&60.82 $\pm$ 3.7 &62.47 $\pm$ 1.5 

% & 65.89 $\pm$ 2.9
% \\

% ~\cite{Ren2020balms} {w/ KD}& 
% 62.75 $\pm$ 3.2&59.27 $\pm$ 8.7&61.01 $\pm$ 3.8 &63.13 $\pm$ 2.7 
% & 0
% \\

% \textbf{Fed-Mas (ours)}&
% 62.08 $\pm$ 2.2&62.06 $\pm$ 7.2&\textbf{62.07 $\pm$ 3.5 }&\textbf{63.32 $\pm$ 2.4}
% & 0
% \\
\hline
  & \multicolumn{5}{c}{With ImageNet Weight Initialization}
    \\
    \hline
FedLC~\cite{pmlr-v162-zhang22p_fedlc} & 71.11 $\pm$ 1.8&73.64 $\pm$ 6.6&72.38 $\pm$ 2.9 &71.63 $\pm$ 1.6 &
76.54 $\pm$ 2.6
\\
BSM-FL~\cite{Ren2020balms} (Baseline) &
73.88 $\pm$ 1.4&74.78 $\pm$ 5.4&74.33 $\pm$ 2.5 &74.49 $\pm$ 1.4 &
78.19 $\pm$ 1.8
\\
~\cite{Ren2020balms} {w/ KD}
&73.87 $\pm$ 1.6&72.44 $\pm$ 5.9&73.16 $\pm$ 3.0 &74.09 $\pm$ 1.5 &
79.17 $\pm$ 2.1
\\
\textbf{Fed-Mas (ours)}
&73.43 $\pm$ 1.6&77.0 $\pm$ 6.6&\textbf{75.21 $\pm$ 2.9} &\textbf{74.61 $\pm$ 1.4} &
\textbf{80.87 $\pm$ 2.2}
\\

% Mas-Inv (ours) &75.02 $\pm$ 1.9&77.06 $\pm$ 6.0&76.04 $\pm$ 3.0 &76.04 $\pm$ 1.9 \\

\bottomrule
  \end{tabular}
  }
    
    % \vspace{-0.5em}
\end{table}

\subsection{Performance on ISIC-FL}
We evaluate the best-performing and competitive methods with the ISIC-FL dataset to shorten the benchmark. While previous studies neglect weight initialization to provide better convergence analysis as pre-trained weights are architecture dependent. Recently,~\cite{nguyen2023where_to_begin} and~\cite{chen2023on_pre_trained_fl_1} studied the impact of pre-training initialization on reducing the data and system heterogeneity in FL. We present in~\Cref{tbl:sota_results_isic_fl} the results of the most competitive methods 
with weight initialization
on the ISIC-FL attribute setting.
% (See~\cref{tblxxx} in Appendix w/o weight initalization).
FedLC~\cite{pmlr-v162-zhang22p_fedlc} demonstrates compelling performance to address label skewness in Hyperkvasir-FL. Nevertheless, it falls short in accommodating attribute heterogeneity in ISIC-FL due to its local learning focus. Our method consistently outperforms FedLC~\cite{pmlr-v162-zhang22p_fedlc} with a notable improvement of 2.8\% and 3.0\% in terms of the averaged balanced accuracies ``Avg'' and balanced accuracy ``B-acc'' respectively. 
% when clients' model weights are not available. 
% When the client's model weights are available and initialized with ImageNet weights, the improvements are x.x\% and x.x\% on the ``Avg'' 
% and ``B-acc'' respectively. 
Compared to the baseline~\cite{Ren2020balms}, Fed-MAS notably captured the underrepresented attribute with 2.2\% on the ``B-acc'' of the ``Dark Attribute'' with a minimal drop of 0.5\% on the ``B-acc'' of the ``Light Attribute'', balancing the intra-class attribute characteristics in FL. On the highly heterogeneous Flamby-ISIC split resembling six hospitals, Fed-MAS outperform FedLC and the baseline on the ``B-acc'' with 4.33\% and 2.68\%, respectively.

% Our method
% when clients' models are not available, and x.x\% and  x.x \% on ``Avg'' and ``B-Acc'' respectively when client's model weights are available. 

\subsection{Privacy Concerns}
Similarly to traditional FL methods~\cite{McMahan2017CommunicationEfficientLO_fedavg,li2021model_moon,LiSZSTS20_fedprox}, Fed-MAS shares the model weights with an additional scalar, $RF$, which protects data privacy by not revealing input data or label distribution. The \textit{scalar}, $RF$, is calculated in the output feature space, safeguarding the input data distribution. Moreover, $RF$ poses uncertainty in approximating the client's label distribution as it can be influenced by diverse attributes in the majority class or a common attribute in the minority class.

\section{Conclusion}
Highly Imbalanced datasets are present in most medical image classifications. This work presents Fed-MAS to deal with this problem. We show that publicly available self-supervised models benefit the FL training procedure more than restricting the optimization direction by incorporating the global attribute imbalance. Future work can explore delayed re-weighting to unleash non-vanishing terms and explore MAS with different local learning strategies in FL settings. 

\vspace{2mm}
\noindent \textbf{Acknowledgement} \small{M.E is partially funded by the EACEA Erasmus Mundus grant. RM is partially funded by the research project PID2021-123390OB-C21 funded by the Spanish Science and Innovation Ministry. This work was supported by the Hong Kong Innovation and Technology Fund under Projects PRP/041/22FX and ITS/030/21, as well as by grants from Foshan HKUST Projects under Grants FSUST21-HKUST10E and FSUST21-HKUST11E.}

\vspace{-1mm}

%
%
% BibTeX users should specify bibliography style 'splncs04'.
% References will then be sorted and formatted in the correct style.
%
% \bibliographystyle{splncs04}
% \bibliography{mybibliography}
%
% \clearpage
\bibliographystyle{splncs04}
\bibliography{egbib}
\clearpage

% This is samplepaper.tex, a sample chapter demonstrating the
% LLNCS macro package for Springer Computer Science proceedings;
% Version 2.21 of 2022/01/12
%
% \documentclass[runningheads]{llncs}
% %
% \usepackage[T1]{fontenc}
% \usepackage{algorithm}
% \usepackage[noend]{algpseudocode}
% \usepackage{multirow}

% % T1 fonts will be used to generate the final print and online PDFs,
% % so please use T1 fonts in your manuscript whenever possible.
% % Other font encondings may result in incorrect characters.
% %
% \usepackage{graphicx}

% \usepackage[capitalize]{cleveref}
% \crefname{section}{Sec.}{Secs.}
% \Crefname{section}{Section}{Sections}
% \Crefname{table}{Table}{Tables}
% \crefname{table}{Tab.}{Tabs.}

% Used for displaying a sample figure. If possible, figure files should
% be included in EPS format.
%
% If you use the hyperref package, please uncomment the following two lines
% to display URLs in blue roman font according to Springer's eBook style:
%\usepackage{color}
%\renewcommand\UrlFont{\color{blue}\rmfamily}
%

% \usepackage{xcolor}

% \usepackage{soul}
% \usepackage{tabularx,booktabs} 
% \usepackage{algorithm}

% \usepackage{xr}
% \usepackage{zref-xr}
% \externaldocument{main}

% \newcolumntype{P}[1]{>{\centering\arraybackslash}p{#1}}
% \newcolumntype{M}[1]{>{\centering\arraybackslash}m{#1}}

\title{Appendix for “Fed-MAS”}

\author{Marawan Elbatel\inst{1,2} \and
Hualiang Wang\inst{1} \and Robert Martí\inst{2} \and 
Huazhu Fu\inst{3}  \and Xiaomeng Li\inst{1} 
}

\authorrunning{Elbatel et al.} 
% % First names are abbreviated in the running head.
% % If there are more than two authors, 'et al.' is used.
% %
\institute{The Hong Kong University of Science and Technology \and Computer Vision and Robotics Institute, University of Girona \and Institute of High Performance Computing (IHPC), Agency for Science, Technology and Research (A*STAR), Singapore}

\maketitle

\appendix
\setcounter{figure}{3}
\setcounter{table}{3}
% Attributes visualization
%Algorithm
%Table with ablation, and two figures of CIFAR-10-
%TSNE visualization
%Table with BSM

\begin{table}
\parbox{.45\linewidth}{
\centering
\caption{HyperKvasir $\lambda_{f}$ ablation.}
    \resizebox{0.5\columnwidth}{!}{
 \begin{tabular}{l|ccc|ccc}
        \toprule
    {\multirow{2}{*}{Method}} &\multicolumn{3}{c|}{IID} & \multicolumn{3}{c}{non-IID}          
    \\
    \cline{2-7}
        & $\lambda_{f}=0$ & $\lambda_{f}=1$ & $\lambda_{f}=3$ & $\lambda_{f}=0$ & $\lambda_{f}=1$ & $\lambda_{f}=3$
        \\
        \toprule
        Fed-MAS & 
        60.28&61.43&\textbf{62.08}&
        59.15& 61.08 & \textbf{61.61}\\
        \bottomrule
    \end{tabular}
    \label{apx:tbl:free_ablation}
    }
    \hfill
    \centering
    \caption{HyperKvasir $f_{\xi}$ ablation non-IID.}
    \resizebox{0.4\columnwidth}{!}{
 \begin{tabular}{l|cc}
\hline
  \multicolumn{1}{l|}{$f_{\xi}$} & All & B-Acc  \\
   \toprule
           CLIP-ViTB/32& 
	60.34 &61.39$\pm$2.1
 \\
        MoCo-RN50 & 
	60.57 &61.61$\pm$1.0
 \\
 \bottomrule
    \end{tabular}
    \label{apx:tbl:ablation_lunch}
    }
}
\hfill
\parbox{0.5\linewidth}{
\centering
\caption{HyperKvasir FL methods with local BRM~\cite{Ren2020balms}.}
    \resizebox{0.5\columnwidth}{!}{
\begin{tabular}{l|cc|cc}
\hline
\multicolumn{1}{c|}{\multirow{2}{*}{Method}}
& \multicolumn{2}{c|}{All}&\multicolumn{2}{c}{B-Acc}\\
\cline{2-5}
&IID & non-IID & IID & non-IID2\\
\hline

FedAvg&58.92	
        &58.24 & 60.28$\pm$0.6&59.15$\pm$1.3 \\

FedProx&59.37
        &58.86&60.47$\pm$1.3&59.64$\pm$2.0\\

Moon &59.45
     &58.72&60.69$\pm$0.9&59.66$\pm$0.8\\

Ours &\textbf{61.05}
&\textbf{60.57}&\textbf{62.08$\pm$0.2}&\textbf{61.61$\pm$1.4} 
   \\
   \bottomrule
\end{tabular}
    \label{apx:tbl:FL_BRM}
    }
      \hfill
    \centering
% <--
\caption{Using a plug-in cRT~\cite{Kang2020Decoupling} on HyperKvasir on non-IID.}
\resizebox{0.3\columnwidth}{!}{
\begin{tabular}{l|cc}
\hline
Method + cRT 
& All 
& \multicolumn{1}{c}{B-Acc}\\ \hline 
Decoupling~\cite{Kang2020Decoupling}
&54.21
&55.6
\\
BSM-FL~\cite{Ren2020balms} 
&62.67
& 63.11
\\
\textbf{Ours} 
&\textbf{65.05}
&\textbf{65.11}
\\

\bottomrule
\end{tabular}
    \label{apx:tbl:FL_CRT}
}

}
\end{table}

\begin{table}[t]
  \centering
  \caption{Flamby-ISIC~\cite{NEURIPS2022_232eee8e_flamby} results on the first fold with the global model (gFL) and the local models (pFL) with ImageNet Weight Initialization. MOON~\cite{li2021model_moon} and FedProx~\cite{LiSZSTS20_fedprox} are reported with local BRM~\cite{Ren2020balms}.}
  \setlength{\tabcolsep}{12pt}
  \label{tbl:sota_results_isic_fl_2}
  \resizebox{0.8\columnwidth}{!}{
  \begin{tabular}{l|cc|l|cc|}
    \hline
    \multirow{2}{*}{Method}
     & \multicolumn{2}{c|}{\textbf{Metric}} 
     &   \multirow{2}{*}{Method}
     & \multicolumn{2}{c|}{\textbf{Metric}}
     \\ \cline{2-3} \cline{5-6}
     & gFL &pFL&&gFL &pFL
     \\
    \hline
MOON ($\mu=0.01$)&72.13&80.03&
FedProx ($\mu=0.1$)& 72.47&79.82 \\

MOON ($\mu=0.1$)&72.45&79.64&
FedProx ($\mu=0.01$)& 73.25 &79.82 \\

MOON ($\mu=1$) &73.12&79.46&
FedProx ($\mu=0.001$)& 73.52 &79.70 \\
FedLC~\cite{pmlr-v162-zhang22p_fedlc} &68.07&78.60&
BSM-FL~\cite{Ren2020balms}& 72.83 &79.79 \\

~\cite{Ren2020balms} w/ KD ($\lambda_{f}$=1)
&72.26&79.66&
~\cite{Ren2020balms} w/ KD  ($\lambda_{f}$=3)
&72.85&80.06\\

\textbf{Fed-MAS} ($\lambda_{f}$=1)
&72.94&82.73&
\textbf{Fed-MAS} ($\lambda_{f}$=3)
&\textbf{74.12}&\textbf{83.28}\\
\bottomrule
  \end{tabular}
  }
    
    % \vspace{-0.5em}
\end{table}

\begin{algorithm}
\centering
\caption{Pseudocode for Fed-MAS.}\label{algo:fedfree}
\begin{algorithmic}[1]
\State \text{\textbf{Notations}} total number of clients (C), server (S), total communication rounds (R), local epochs (E), learning rate ($\eta$), and a set of client’s data sliced into batches of size B ($\mathcal{B}$).
\State \text{\textbf{\underline{ServerExecution:}}}
\State Init $\theta_{glob}^{1}$
\For{\textit{each round} $r = 1, ..., R$}
  % \State $S_t \gets$ (Selection of c clients from C)\;
  \For{\textit{client} $c \in C$ \textit{in parallel}}
      \State $\theta_{c}, {RF}_{c}\gets$\textbf{LocalUpdate}($\theta_{glob}^r$);\;
\EndFor
  % \State $\bar{RF}_c \gets {\frac{RF_{c}}{\sum\limits_{j}RF_{j}}}$ \;
  \State ${\theta^{r+1}_{glob}\gets}$\textbf{DLMA}($RF_{c},\theta'_{c}$,c = 1 to C); //~\cref{eq:tempavg}\; 
   \EndFor
\State \textbf{Return}  $\theta_{glob}^{R}$
%   \State 
\State \underline{\textbf{LocalUpdate}} ($\theta_{glob}$): 
    \State Init $\hat{w}_{k}=0;$\;
    \State Init $ w_k  =\mathcal{L}_{k}^{\theta_{glob}};$ \;
    % following \cref{eq:global_measurement}
    \For{\textit{each local epoch} $e = 1, ..., E$}
        \For{\textit{each batch b} $\in\mathcal{B}$}
             \State $\mathcal{L}_{total} =\mathcal{L}_{sup}+ \lambda_{f} \mathcal{L}_{f};$~//~\cref{eq:loss_total}\;
             
            \State $\theta' \leftarrow \theta' - \eta \triangledown \mathcal{L}_{total};$ \;
            \State $\hat{w}_{k} \leftarrow  \hat{w}_{k} + \mathcal{L}_f(b_k);$ // running distillation loss mean for each class k\; 
            %~\cref{eq:local_estimation} 
        \EndFor
    \EndFor
\State $RF=\sum\limits_{k=1}^{K}  w_{k} \hat{w}_{k};$ \; 
// RF~$\uparrow \approx$ divergence~$\theta_{glob},f_{\xi}\uparrow$
\State \textbf{Return} $\theta',RF$
\end{algorithmic}
\end{algorithm}

\begin{figure}[!htb]
\centering
    \begin{tabular}{c c}
    \large{$w/o~MLP$} & \large{$MLP$}
        \\
        \includegraphics[width=.36\textwidth]{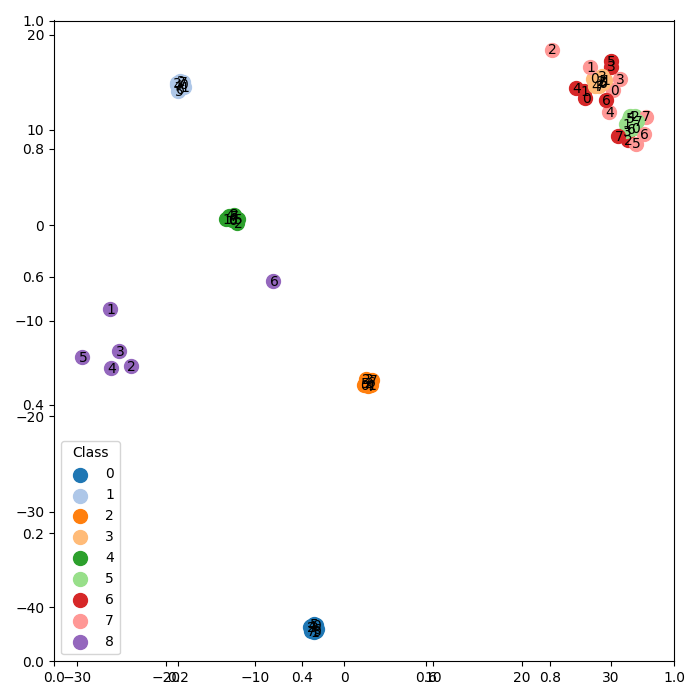}&
        \includegraphics[width=.36\textwidth]{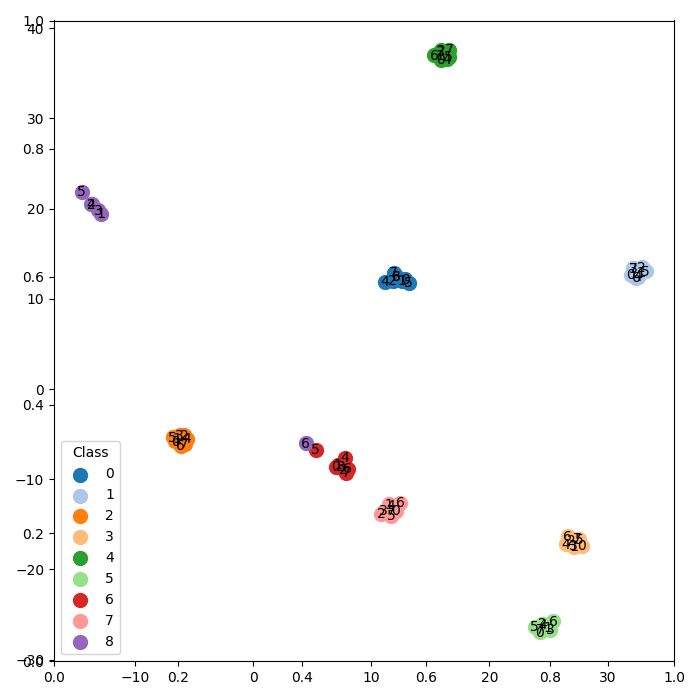}\\
    \end{tabular}
\caption{Feature representation with and without the learnable projector $MLP_{\theta}$. We sample a subset of head (0,1,2), medium (3,4,5), and tail (6,7,8) classes for feature visualization across different clients. Each point represents the mean feature output for each class (color) in each client (point).}
\label{Apx:fig:TSNE}
\end{figure}

\begin{figure}[h]
\centering
    \begin{tabular}{c c}
    \textit{Accuracy} & \textit{ $\mathcal {L}_{sup}$/$\mathcal {L}_{f}$ }
        \\
        \includegraphics[width=.44\textwidth]{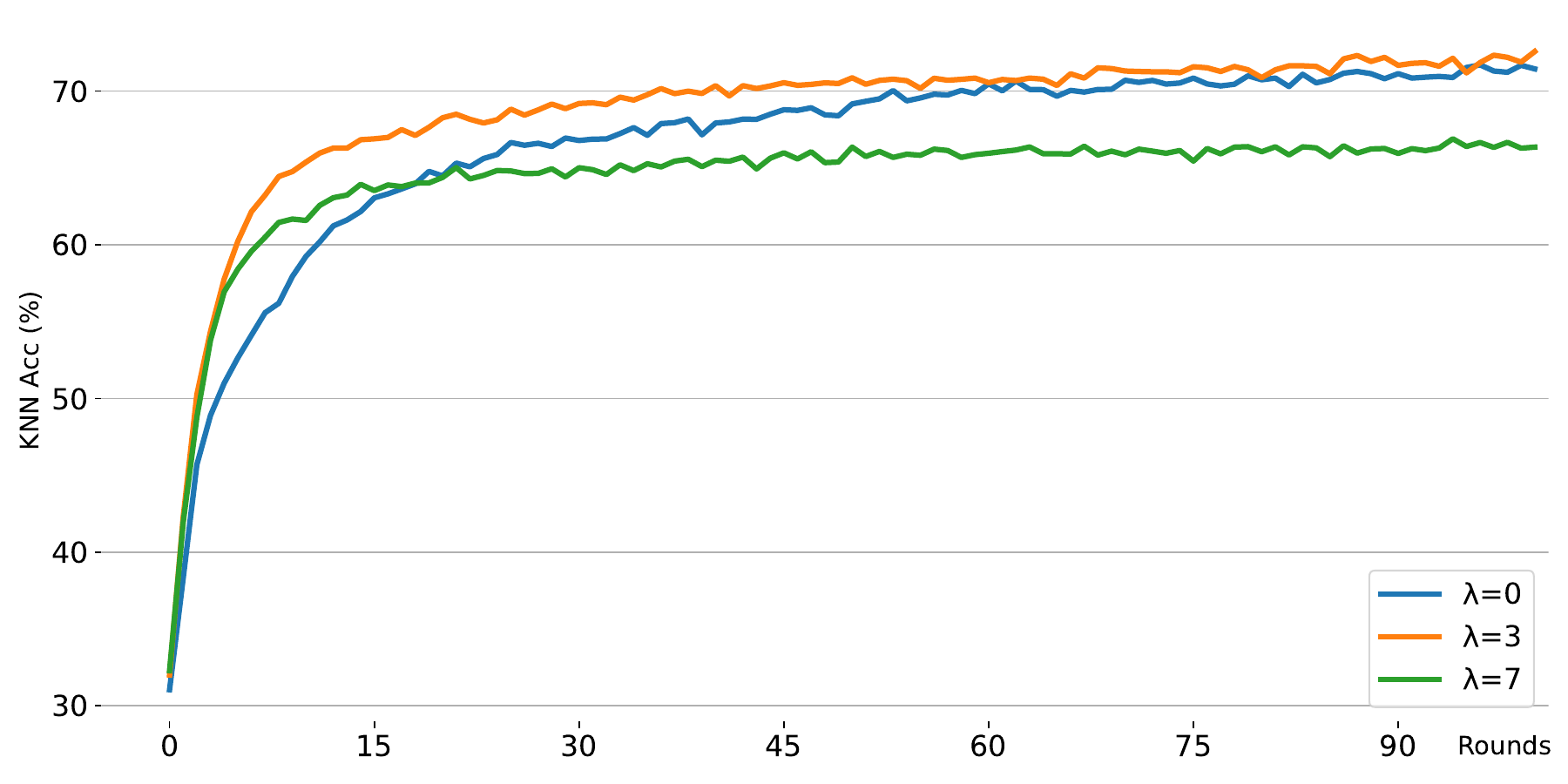}&
        \includegraphics[width=.44\textwidth]{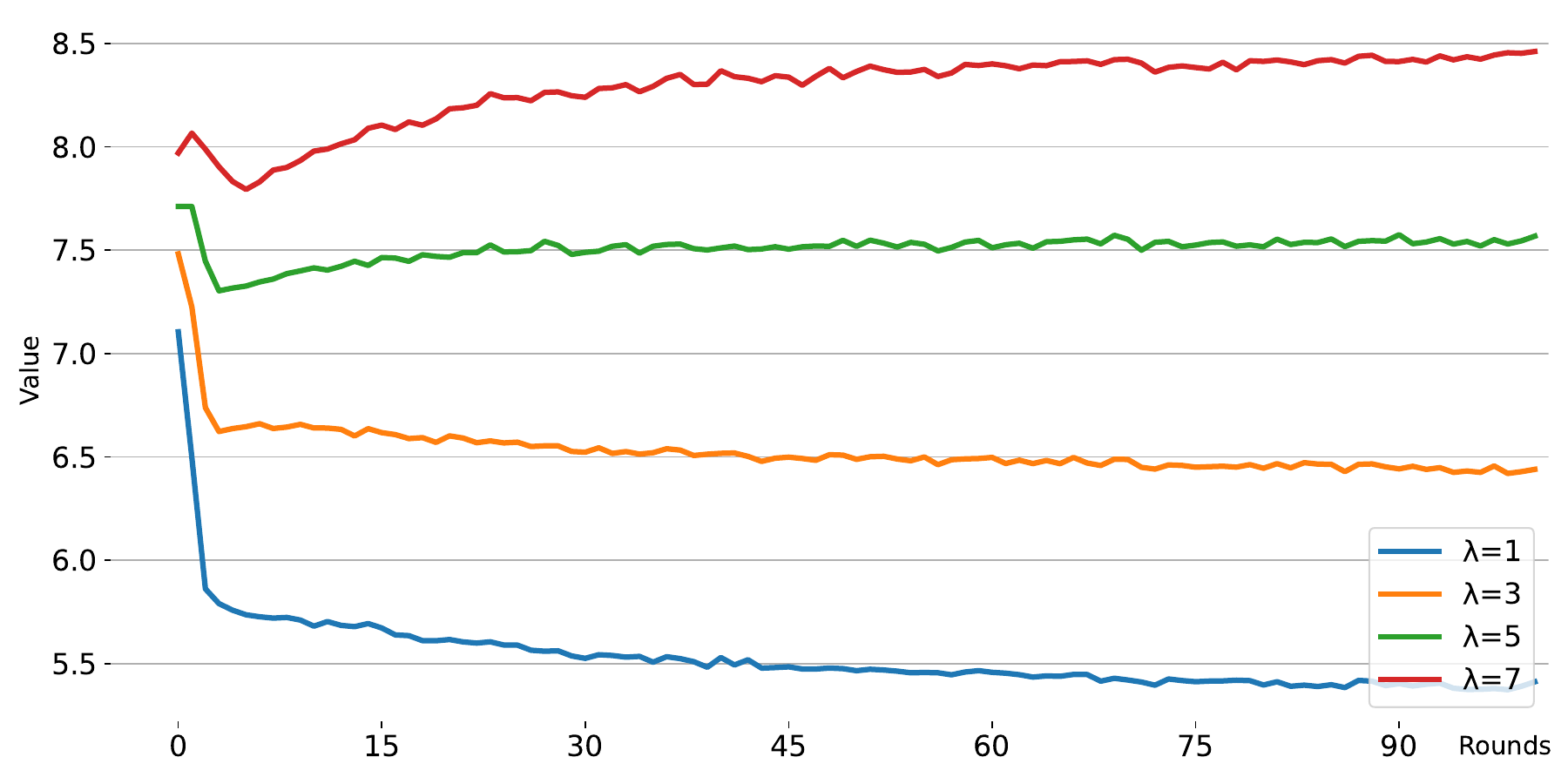}\\
    \end{tabular}
\caption{Higher Value of $\lambda_{f}$ ($\lambda_{f}=7$) causes task deviation. $\lambda=3$ show faster convergence (Acc.), and  make $L_{sup}/L_{f}$ ratio consistent on a toy dataset (CIFAR-100 non-iid).}
\label{Apx:fig:lambda_ablation}
\end{figure}

\clearpage

\end{document}